\documentclass[runningheads]{llncs}
\PassOptionsToPackage{prologue,dvipsnames,table}{xcolor}

\usepackage[mobile]{eccv}
\usepackage{scontents}
\usepackage{multicol}
\usepackage{multirow}
\usepackage{float}
\usepackage{caption}
\usepackage{overpic}

\usepackage{enumitem}
\newlist{compactitem}{itemize}{1}
\setlist[compactitem]{topsep=0pt,partopsep=0pt,itemsep=0pt,parsep=0pt,leftmargin=*}
\setlist[compactitem,1]{label=\textbullet}

\usepackage[dvipsnames]{xcolor}

\definecolor{turquoise}{cmyk}{0.65,0,0.1,0.3}
\definecolor{purple}{rgb}{0.65,0,0.65}
\definecolor{dark_green}{rgb}{0, 0.5, 0}
\definecolor{orange}{rgb}{0.8, 0.6, 0.2}
\definecolor{red}{rgb}{0.8, 0.2, 0.2}
\definecolor{darkred}{rgb}{0.6, 0.1, 0.05}
\definecolor{blueish}{rgb}{0.0, 0.3, .6}
\definecolor{light_gray}{rgb}{0.7, 0.7, .7}
\definecolor{pink}{rgb}{0.9, 0, 0.6}
\definecolor{greyblue}{rgb}{0.25, 0.25, 1}
\definecolor{teal}{rgb}{0.0, 0.4, 0.4}
\definecolor{chocolate}{rgb}{1.0, 0.4, 0.0}

\definecolor{1st}{rgb}{0.9, 0.65, 0.65}
\definecolor{2nd}{rgb}{1, 0.9, 0.9}

\renewcommand{\paragraph}[1]{\vspace{.5em}\noindent\textbf{#1}}

\DeclareMathOperator*{\argmax}{arg\,max}

\AtEndPreamble{
\usepackage[capitalize]{cleveref}
\crefname{section}{Section}{Sections}
\Crefname{section}{Section}{Sections}
\crefname{table}{Table}{Tables}
\Crefname{table}{Table}{Tables}
\Crefname{equation}{Eq.}{Eqs.}
\crefname{equation}{}{}
}

\usepackage{soul}
\setuldepth{foobar}

\newcommand{\real}{\mathbb{R}}

\newcommand{\f}{\mathbf{f}}

\newcommand{\x}{\mathbf{x}}

\newcommand{\fielddim}{D}
\newcommand{\gridres}{N}
\newcommand{\gridVertex}{\mathbf{v}}
\newcommand{\mean}{\boldsymbol{\mu}}

\newcommand{\interpWeights}{\alpha}
\newcommand{\nVertices}{V}
\newcommand{\hash}{\mathcal{H}}
\newcommand{\pars}{\boldsymbol{\theta}}

\newcommand{\concat}{\ensuremath{\oplus}}
\newcommand{\InstantNGP}{InstantNGP~\cite{muller2022instant}\xspace}
\newcommand{\field}{\mathcal{F}}
\newcommand{\nLevels}{L}
\newcommand{\iLevels}{l}
\newcommand{\nInjective}{B}
\newcommand{\F}{\mathbf{F}}
\newcommand{\loss}[1]{\mathcal{L}_\text{#1}}

\usepackage{eccvabbrv}

\usepackage{graphicx}
\usepackage{booktabs}

\usepackage[accsupp]{axessibility}  

\usepackage[breaklinks,colorlinks,citecolor=eccvblue]{hyperref}

\usepackage{orcidlink}

\usepackage{comment}

\usepackage{wrapfig}

\makeatletter
\newcommand{\printfnsymbol}[1]{%
        \textsuperscript{\@fnsymbol{#1}}%
}
\makeatother

\begin{document}

\title{Lagrangian Hashing \\ for Compressed Neural Field Representations}

\titlerunning{Lagrangian Hashing}

\author{Shrisudhan Govindarajan\thanks{Authors contributed equally to this work.} \inst{1}\orcidlink{0000-0002-3546-8223} \and %
Zeno Sambugaro\printfnsymbol{1} \inst{2}\orcidlink{0000-0002-4541-4528} \and
Akhmedkhan (Ahan) Shabanov\inst{1}\orcidlink{0009-0009-3526-4640} \and
Towaki Takikawa\inst{3}\orcidlink{0000-0003-2019-1564} \and
Daniel Rebain\inst{4}\orcidlink{0000-0003-4691-7909} \and
Weiwei Sun\inst{4} \orcidlink{0000-0002-7640-0006}\and
Nicola Conci\inst{2}\orcidlink{0000-0002-7858-0928} \and
Kwang Moo Yi\inst{4}\orcidlink{0000-0001-9036-3822} \and
Andrea Tagliasacchi\inst{1,3,5}\orcidlink{0000-0002-2209-7187}}

\authorrunning{S.~Govindarajan, Z.Sambugaro et al.}

\institute{\textsuperscript{1}Simon Fraser University, \textsuperscript{2}University of Trento, \textsuperscript{3}University of Toronto, \\ \textsuperscript{4}University of British Columbia, \textsuperscript{5}Google DeepMind}

\maketitle
\begin{center}
    Project page: \url{https://theialab.github.io/laghashes/}
\end{center}

\begin{figure}
    \centering
    \includegraphics[width=\columnwidth]{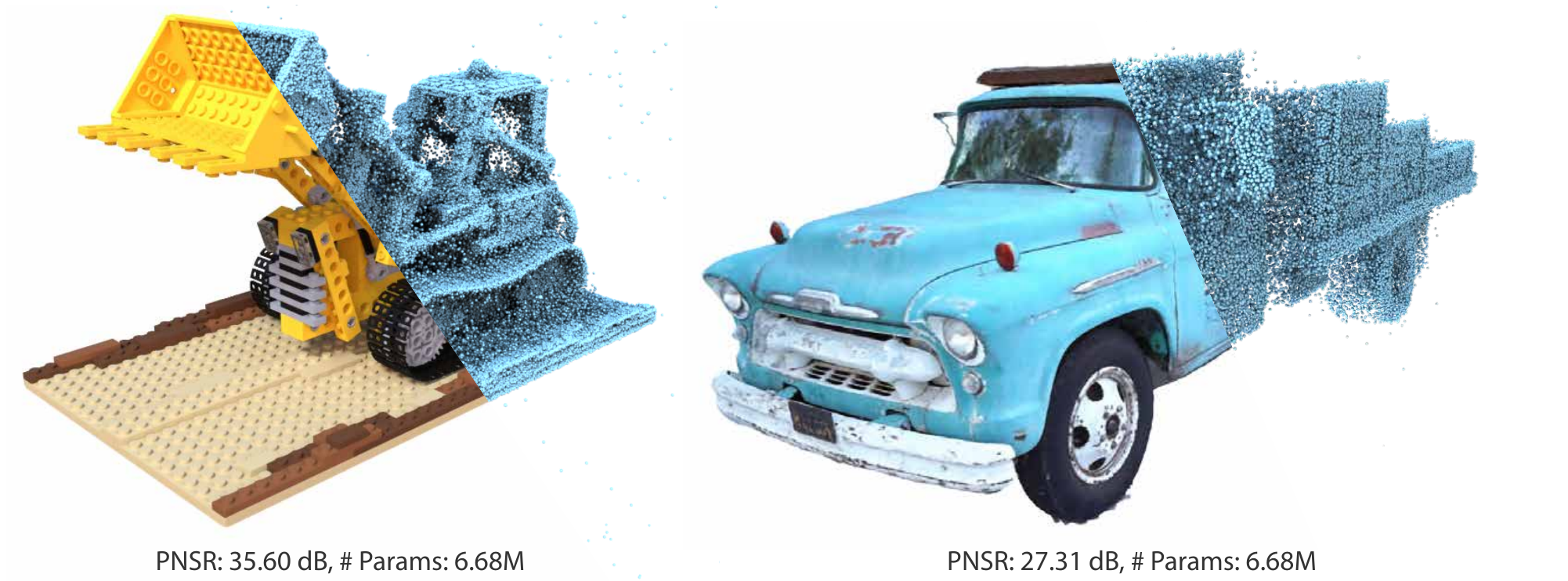}
    \caption{We introduce a hybrid representation that is simultaneously Eulerian (grids) and Lagrangian (points), which realizes high-quality novel view synthesis as shown above, while at the same time being more memory efficient.
    }
    \label{fig:teaser}
\end{figure}
\vspace{-3.0em}
\begin{abstract}
We present Lagrangian Hashing, a representation for neural fields combining the characteristics of fast training NeRF methods that rely on Eulerian grids (i.e.~InstantNGP), with those that employ points equipped with features as a way to represent information (e.g. 3D Gaussian Splatting or PointNeRF).
We achieve this by incorporating a point-based representation into the high-resolution layers of the hierarchical hash tables of an InstantNGP representation.
As our points are equipped with a field of influence, our representation can be interpreted as a mixture of Gaussians stored within the hash table.
We propose a loss that encourages the movement of our Gaussians towards regions that require more representation budget to be sufficiently well represented.
Our main finding is that our representation allows the reconstruction of signals using a more compact representation without compromising quality.
\keywords{Lagrangian \and Point Cloud \and Compact Representation \and Neural Field}
\end{abstract}

\section{Introduction}
\label{sec:intro}

As immersive mixed reality interfaces and volumetric content capture systems become popular, there is an increasing demand for a multimedia format that can compactly represent and transmit various types of multimedia. The diversity of multimedia formats (images, videos, volumetric 3D, radiance fields, etc) in mixed reality systems necessitates the codecs themselves to also be flexible to handle different types of formats. 

Neural fields~\cite{neuralfields2021} have emerged as a general data format that can represent different multimedia formats with a unified codec based on model fitting. They have been popular in recent literature for representing all sorts of data, but in particular have been widely used for radiance field reconstruction~\cite{mildenhall2021nerf}.
In contrast to traditional multimedia formats that convert data into alternate formats through transformations, neural fields convert data by fitting a model to the data via optimization.
This adds to the flexibility of the transformation by allowing the integration of additional objective functions and scenarios, like 3D reconstruction in an inverse problem setting.

A class of neural field models that have been particularly successful are \textit{feature grids}, which uses a differentiable data structure holding features and a small multi-layer perceptron (MLP) as the model.
These models inherit the strengths of highly performant data structure (such as an octree~\cite{takikawa2021neural} or hash grid~\cite{muller2022instant}) which allow the neural fields to fit complex data with large spatial extent without sacrificing performance.
These feature grids methods, however, typically come at the cost of a larger memory footprint.

Although many data structure tricks like sparsity~\cite{takikawa2021neural, martel2021acorn}, low-rank factorization~\cite{chen2022tensorf}, linear transforms~\cite{rho2022masked}, and hash probing~\cite{takikawa2023compact} have been proposed to improve the \textit{memory-quality tradeoff curve} of feature grids, these works do not fundamentally address the spatially non-homogeneous structure of 3D data: 
more features should be allocated for parts of the data with higher complexity.
Achieving this objective would let the representation use the available memory footprint more efficiently.

These feature grid-based neural fields typically represent data in an \textit{Eulerian} way, as a vector field over some coordinate system.
Even if they use sparse or factorized data structures, they generally use a grid where vertices are laid out in uniform intervals which allows for simple implementations of indexing algorithms. \textit{Lagrangian} ways of representing data, on the other hand, would allow the representation to flexibly allocate the grid points in space, but may suffer from more complex indexing algorithms that may involve techniques like approximate nearest neighbour searches.
\footnote{We refer to Eulerian and Lagrangian representations in numerical physics, where
one can store the state of the system (\eg, velocity of a fluid) on either grids (Eulerian) or on particles (Lagrangian).
}

Our work, Lagrangian Hashing (LagHash), marries the simplicity and performance of \textit{Eulerian} representations where features are laid out on uniform grids with a \textit{Lagrangian} representation that employs a point-based representation in which features can freely move around in space.
In more details, our representation builds upon hierarchical hashes introduced by \InstantNGP, but, in each hash bucket, rather than just storing a feature, we store a small set of points, each equipped with a feature.\footnote{Note this is done on a selection of levels, and therefore our representation is an Eulerian-Lagrangian hybrid.} 
While other point-based representations require acceleration data structures to access the points, our representation \textit{reuses} the hash implementation.
Most importantly, as point-clouds can adaptively allocate representation budget by increasing the density of the point cloud where needed, they are a more effective representation for storing high-frequency information, which results in smaller models that achieve similar visual performance.
\section{Related Works}
\label{sec:related}
In what follows, we review the literature on traditional compression, neural compression method and point-based representation.

\paragraph{Traditional Compression.} The dominant approach in lossy compression for traditional multimedia usually involve transform coding~\cite{goyal2001theoretical}, quantization~\cite{gray1998quantization}, and entropy coding~\cite{huffman1952method}. Linear projections into a fixed basis space (like discrete cosine transform~\cite{ahmed1974discrete}) are often used in practice for image and video codecs like JPEG~\cite{wallac1991jpeg}. Transform coding methods usually treat the multimedia data as a collection of vectors, where each vector represents a local fixed-size patch of pixels. These \textit{block-based} methods also exist for 3D volumetric data~\cite{balsa2014state}, where 3D patches of voxels are encoded using linear transformations~\cite{de2016compression, tang2018real, tang2020deep}. These methods are not spatially adaptive, in that each equal-sized patch of an image or volume go through the same encoding regardless of the local resolution content. In contrast, our work uses a point-based, Lagrangian formulation where more feature vectors can be placed where there is more complexity of data. The spatial adaptivity allows our work to more efficiently allocate resources. 

\paragraph{Neural Compression.} Instead of transforming data, model-based compression methods compress data by fitting a model to the data. These models can be polynomials~\cite{eden1986polynomial}, partial differential equations~\cite{galic2008image}, gaussian mixture models~\cite{cheng2020learned}, or neural fields~\cite{song2015vector}. Works that use neural fields either use multi-layer perceptrons~\cite{song2015vector, dupont2021coin, strumpler2022implicit} or feature grids~\cite{muller2022instant, takikawa2022variable, takikawa2023compact}. Most compression methods that use neural fields have still not adapted models that use point-based representations which have an advantage of spatial adaptivity. In contrast, our work uses a hybrid of Eulerian feature grids and a Lagrangian representation that enables spatial adaptivity.

\paragraph{Point-based Representations}. 
Point-based representations have been widely explored as a representation in computer graphics\cite{alexa2004point, gross2011point, sridhar2014real, kobbelt2004survey}.
Polygon meshes, which are a dominant representation in computer graphics, can be seen as a form of a point-based representation, as the points lying on a polygonal primitive are represented as convex combinations of its vertices.
Point-based representations have also been used in reconstruction settings in conjunction with neural fields, 
but they frequently either require careful initialization, such as points from depth~\cite{aliev2020neural, kopanas2021point, rakhimov2022npbg++, ost2022neural, xu2022point}, COLMAP point cloud~\cite{kerbl20233d} and LiDAR~\cite{chang2023neural,sun2023pointnerf++}.
In contrast, our method does not require careful initialization, and guides point placement via a carefully designed guidance loss. 

Recently, point-based neural rendering~\cite{lassner2021pulsar, zhang2022differentiable, zhang2023frequency, kerbl20233d} enabled the rendering of 3D point clouds onto images via differentiable rasterization. 
Particularly, Gaussian Splatting~\cite{kerbl20233d} and its follow-ups~\cite{yang2023deformable, yu2023mip, yan2023multi, wu20234d} have achieved impressive results in rendering quality and test-time efficiency. 
Despite their quick adoption, these methods still rely on COLMAP for point initialization, and carefully tuned heuristics to grow and prune points.
More importantly, these methods typically require \textit{large storage}, with millions of points needed to accurately represent a scene, whereas our method naturally leads to a compact representation. 
For more details of recent advances on Gaussian Splatting, we refer the interested readers to a recent survey~\cite{chen2024survey}.

\begin{figure}[!t]
\begin{center}
\begin{overpic}[width=\columnwidth]{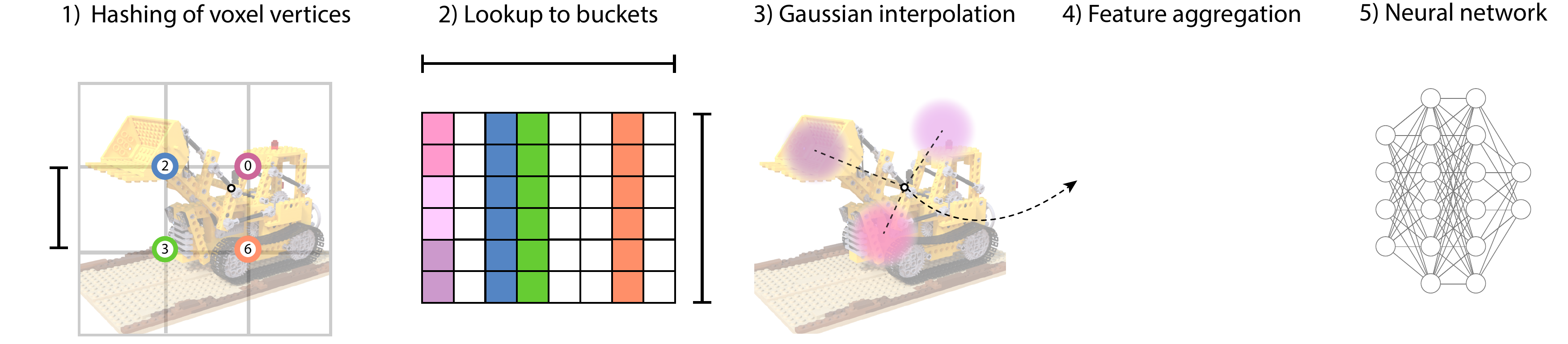}

\put(0, 8){{\small $\frac{1}{N}$}}

\put(15, 8){$x_i$}
\put(34, 18.25){\scalebox{.75}{$B$}}
\put(45, 8){\scalebox{.75}{$K$}}

\put(27.5, 15.5){{\tiny $0$}}
\put(29.5, 15.5){{\tiny $1$}}
\put(31.5, 15.5){{\tiny $2$}}
\put(33.5, 15.5){{\tiny $3$}}
\put(35.5, 15.5){{\tiny $4$}}
\put(37.5, 15.5){{\tiny $5$}}
\put(39.5, 15.5){{\tiny $6$}}
\put(41.5, 15.5){{\tiny $7$}}

\put(24, 13.5){{\tiny $f_1$}}
\put(24, 11.5){{\tiny $\mu_1$}}
\put(24, 9.5){{\tiny $f_2$}}
\put(24, 7.5){{\tiny $\mu_2$}}
\put(24, 5.5){{\tiny $f_3$}}
\put(24, 3.5){{\tiny $\mu_3$}}

\put(58, 14){\scalebox{.75}{$\mu_1,\sigma_1$}}
\put(49.5, 13){\scalebox{.75}{$\mu_2,\sigma_2$}}
\put(54, 6){\scalebox{.75}{$\mu_3,\sigma_3$}}

\put(66, 11){\scalebox{.75}{$\F(x_i) = \sum_k \mathcal{N}_k(x_i) \cdot f_k$}}

\put(98, 11){\scalebox{0.75}{$c(x_i)$}}
\put(98, 8){\scalebox{0.75}{$\tau(x_i)$}}

\end{overpic}
\end{center}
\caption{
\textit{(1) Hashing of voxel vertices:} For any given input coordinate $x_i$, our method identifies surrounding voxels across $L$ Levels of detail (Lods) (Only one Lod is showed for convenience). Indices are then assigned to the vertices of these voxels, through hashing procedure. \textit{(2) Lookup to buckets:} for all resulting corner indices, we look up the corresponding $B$ buckets, containing $K$ feature vector and their corresponding $\mu_{k}$ position. \textit{(3) Gaussian interpolation:} We compute Gaussian weights with respect to the input position for every feature vector in the bucket. \textit{(4) Feature aggregation:} We multiply the Gaussian weights for the feature corresponding to the feature vector and aggregate them from every level of detail. \textit{(5) Neural Network}: the resulting concatenated features are mapped to the input domain by the Neural Network.
}
\label{fig:overview}
\end{figure}
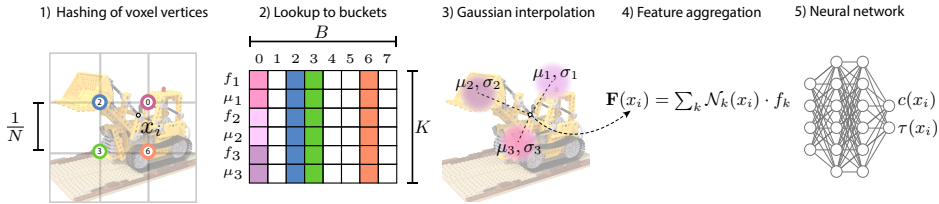
\section{Method}
\label{sec:method}
We introduce a new representation that combines the Eulerian nature of fast-training NeRF methods~\cite{muller2022instant} to the Lagrangian nature of emergent 3D Gaussian Splatting (3DGS) representations~\cite{kerbl20233d}.
Building directly on top of the hierarchical Eulerian representation introduced by \InstantNGP, we achieve this by incorporating a point-based representation in the high-resolution layers of its hash tables.
We select the high-resolution layers as to let the model focus its representation power to precise locations in space, akin to how 3DGS~\cite{kerbl20233d} employs small Gaussians to capture fine-grain details of the scene.
We (implicitly) equip each point with a standard deviation proportional to the grid resolution, hence our representation can be interpreted as a mixture of Gaussian with non-trainable standard deviation and mixture weights.
Differently from 3DGS, we employ \textit{isotropic} Gaussians, and the standard deviation associated with each point describes the portion of space that the feature stored alongside the point position is meant to represent.

\paragraph{Overview}
A visual overview of our representation can be found in~\Cref{fig:overview}.
We reviewing the hashed multi-scale representation in~\cref{sec:multilevel}, how features are interpolated within each level in~\cref{sec:perlevel}, and then detail how to augment hash buckets with a mixture-of-Gaussians representation in~\cref{sec:bucket}.
We discuss our training methodology in~\cref{sec:losses}, including the introduction of a loss that is critical to guide the MoG towards regions of space that require additional representation power.

\subsection{Multi-scale representation}
\label{sec:multilevel}
Analogously to \InstantNGP, our architecture produces a field value $\field(\x)$ at an arbitrary position in space $\x$ by interpolating feature vectors evaluated at the vertices of a stack $\iLevels=\{1, \dots, \nLevels\}$ of regular grids.
Their resolution follows the geometric progression $\gridres_\iLevels=\gridres_\text{min} \cdot b^\iLevels$, where $\nLevels$, $\gridres_\text{min}$ and $b$ are hyper-parameters.
The field value is computed by concatenating ($\oplus$) the features across levels, and then passing this vector through a shared decoder with parameters~$\pars$:
\begin{equation}
\field(\x) = \text{MLP}(\f_1(\x) \concat \f_2(\x) \concat ... \concat \f_L(\x); \pars), 
\quad
\quad
\field(\x): \real^\fielddim \rightarrow \real^F
\end{equation}

\subsection{Per-level feature -- $\f_\iLevels(\x)$}
\label{sec:perlevel}
Let us now consider how each feature $\f_\iLevels(\x)$ is computed.
Denote with $\F$ a tensor of features in memory, and let $\hash(\x)$ be an indexing function that retrieves the \textit{indices} of $\F$ corresponding to the features of the grid corners $\{\gridVertex_v\}$ of field query position $\x$.
Denote the corresponding grid interpolation weights as $\{\interpWeights_v\}$, where $v=\{1, \dots, \nVertices{=}2^\fielddim \}$.
We interpolate features at position $\x$ as:
\begin{equation}
\f_l(\x) = \sum_{v \in \hash_l(\x)} \interpWeights_v \cdot \F_l[v](\x)
\label{eq:interp}
\end{equation}
Similarly to \InstantNGP, we implement $\hash_l$ as an injective map whenever~$\gridres_l<\nInjective$, and as a hash function otherwise.
In other words, $\hash_l: \real^\fielddim \rightarrow [1, \gridres_l-1]$ if $\gridres_l<\nInjective$, and $\hash_l: \real^\fielddim \rightarrow [0, \nInjective-1]$ otherwise, and respectively the feature tensor $\F \in \real^{\gridres_l \times F}$ or $\F \in \real^{\nInjective \times F}$.
As we closely follow \InstantNGP to enable fair comparisons, we refer the reader to this paper for additional details regarding the implementation of $\hash$.
As at finer levels $v$ indexes the buckets of a hashing operation, we refer to $\F_l[v]$ as a ``per-bucket'' feature.

\newcommand{\cutoffLevel}{\tilde{\nLevels}}
\paragraph{Eulerian vs. Lagrangian features.}
Note that, differently from \InstantNGP, the notation in \cref{eq:interp} hints at the fact that the feature grid is \textit{evaluated} at the position $\x$.
The feature evaluation depends on the depth of the corresponding level.
Consider a hierarchical hash of $\nLevels$ levels where the last $\cutoffLevel$ are Lagrangian.
At shallow levels, that is $\iLevels{<}($\nLevels${-}\cutoffLevel$), the feature $\F$ follows \InstantNGP, that is, the feature is retrieved from the hashed Eulerian grid.
In other words, $\F_l[v](\x) \equiv \F_l[v]$.
At deeper levels, we retrieve the features from a Lagrangian representation that takes the form of a \textit{Gaussian mixture}, which we will detail next in \cref{sec:bucket}.

\subsection{Per-bucket feature (Lagrangian) -- $\F_l[v](\x)$}
\label{sec:bucket}
To simplify notation, and without loss of generality, let us drop the bucket index~$[v]$ and the level index $l$.
Within each bucket, we store an isotropic Gaussian mixture consisting of $k=\{1, \dots, K\}$ elements, parameterized by mean $\mean_k$, standard deviation $\sigma_k^2$, and corresponding feature vector $\f_k$.
The feature is computed by evaluating the Gaussian mixture at position $\x$:
\begin{equation}
\F(\x) = \sum_k \mathcal{N}_k(\x) \cdot \f_k, 
\quad
\quad
\mathcal{N}_k(\x) = \frac{1}{(2\pi)^{1/2}\sigma_k}\text{exp}\left(-\frac{\| \x - \mean_{k} \|_2^2}{2\sigma_k^2}\right).
\end{equation}
 
\subsection{Training (novel view synthesis)}
\label{sec:losses}
%
During training, we jointly optimize the shared decoder parameters, the Eulerian representation for levels $\iLevels{<}($\nLevels${-}\cutoffLevel$), and the Lagrangian representation for the last $\cutoffLevel$ levels.
The latter includes the mean and feature, and (optionally) the standard deviation of each Gaussian.
To train our representation, we optimize the loss:
\begin{equation}
\mathcal{L} =
\mathcal{L}_{\text{recon}} +
\lambda_\text{dist} \mathcal{L}_{\text{dist}} + 
\lambda_\text{guide} \mathcal{L}_{\text{guide}}
\end{equation}
where $\mathcal{L}_{\text{recon}}$ is the pixel reconstruction loss computed by volume rendering~\cite{mildenhall2021nerf}, evaluated via a Huber loss analogously to \InstantNGP, $\mathcal{L}_{\text{dist}}$ is the distortion loss proposed in~\cite{barron2022mipnerf360} to promote the formation of surfaces within the volume, and $\mathcal{L}_{\text{guide}}$ avoids vanishing gradients in the optimization of the Lagrangian portion of our representation by guiding the movement of Gaussians towards surfaces.

\paragraph{Guidance loss.}
During training, whenever we back-propagate a position $\x$ with respect to a Gaussian whose mean $\mean$ that is too far from $\x$ (scaled by the standard deviation $\sigma^2$), the computed gradients become very small, which interferes with effective optimization of our representation.
Note that a very similar problem is found in the training of Gaussian Mixture Models, for which Expectation-Maximization (EM) is typically employed to address this issue~\cite{stauffer1999adaptive}.
In defining our loss, we take \textit{inspiration} from EM training of GMMs.
For the moment being, consider having Gaussians at a \textit{single} level, as we will extend this later to the multi-level setting.
In the \textit{E-step}, given a query point $\x$ we identify the Gaussian (across buckets \textit{and} Gaussians therein) whose PDF (scaled by the mixture weights) is maximal:
\begin{equation}
G(\x) = \interpWeights_{v^*} \cdot \mathcal{N}(\x; \mean_{k^*, v^*}, \sigma^2_{k^*, v^*}), 
\quad
k^*, v^* = \argmax_{k,v} \:\: \interpWeights_v \cdot \mathcal{N}_{k,v}(\x)
\end{equation}
Then, in the \textit{M-step}, we optimize the parameters so to minimize the discrepancy between $G(\x)$ and the NeRF integration weights~$W(\x)=T(\x)\cdot \tau(\x)$, which is a PDF along the ray, as derived in~\cite[Eq.~(29)]{nerfdigest}.
\footnote{We denote density with $\tau(\x)$ to avoid confusion with the Gaussian's variance $\sigma^2$.}
We measure the discrepancy between the two PDFs via the KL-divergence, and after dropping the subscripts $(k^*, v^*)$ to simplify notation we note that the equality:
\begin{equation}
\argmax_{\mean, \sigma^2} \:\: 
\text{KL}( W(\x) || G(\x) )
\equiv 
\argmax_{\mean, \sigma^2} \:\: 
- W(\x) \cdot \log(G(\x))
\end{equation}
holds due to the definition of KL divergence, and that the term $W(\x) \cdot \log(W(\x))$ is constant with respect to the Gaussian means, and hence can be dropped.
Similarly to the EM algorithm, our approach involves alternating between two key steps: 1) determining the posterior distribution of latent variables by assigning each point to a Gaussian, and 2) maximizing the KL divergence based on the defined correspondence.
This two-step process can be written as a loss:
\footnote{We draw \textit{inspiration} from EM algorithm, but there are distinctions. 
Our implementation does not incorporate optimizable mixture weights, and we employ a \textit{hard} assignment between points and Gaussians. 
Finally, each point in our model is weighted, contrasting with the EM algorithm in GMM where all points have equal weights.}
\begin{equation}
\mathcal{L}_{\text{guide}}^\iLevels(\x) = - W(\x) \cdot \: \log
\left(
\max_{k,v} \: \interpWeights_{v,l} \cdot \mathcal{N}_{k,v,l}(\x)
\right)
\end{equation}
and as the max commutes with the (monotonic) log operator, after some straightforward algebraic manipulations, we define our loss to its final form:
\begin{equation}
\mathcal{L}_{\text{guide}}(\x) = \sum_\iLevels \mathcal{L}_{\text{guide}}^\iLevels(\x),
\mathcal{L}_{\text{guide}}^\iLevels(\x) = W(\x) \cdot
\min_{k,v}\left(-\log\left(\interpWeights_{v,l}\right) + \frac{\|\x-\mean_{k,v,l}\|_2^2}{2\sigma_{k,v,l}^2}\right)
\label{eq:guide}
\end{equation}
Note that this loss has a very intuitive interpretation.
It states that, while integrating along a ray, if we find a position~$\x$ that is likely to lie on a surface~(i.e. $W(\x) \approx 1$), then there should be \textit{one} Gaussian nearby~(i.e. $\mean \approx \x$).
Further, if we assume constant $\interpWeights$ and $\sigma^2$, note that amongst all possible Gaussians the loss will select the \textit{closest} Gaussian. 
In other words, our loss is a (scaled) one-sided \textit{Chamfer} loss between Eulerian and Lagrangian representations.

\subsection{Implementation}
We now provide an overview of the key implementation details of our method.
\footnote{We provide a copy of our source code in the supplementary materials.}
We based our NeRF implementation on the \textit{nerfacc} framework~\cite{li2023nerfacc} and our 2D image fitting on the Kaolin-wisp library~\cite{KaolinWispLibrary}.
For our experiments, we chose a feature dimension, $F{=}2$, and $L{=}16$ levels of detail (LoDs) for all the experiments in the paper as proposed in InstantNGP.
As the decoder in our architecture, we employ a Multi-layer Perceptron (MLP) with one hidden layer, containing 64 neurons for all the tasks.

For initialization of our decoder, we employ a classical Xavier uniform distribution~\cite{glorot2010understanding}.
The hash table features are initialized using a normal distribution with zero mean and standard deviation of $1e^{-3}$.
We parameterize the Gaussians with a learnable mean and a \textit{fixed} standard deviation.
The initialization of Gaussian means for the image tasks is randomized within the image space, while for Neural Radiance Field (NeRF) reconstruction, they are uniformly distributed within a sphere of radius $0.75$.
In our framework, the standard deviation of each Gaussian defines a field of influence for each feature vector.
The standard deviations are directly related to the spatial domain of the grid vertices.
We initialize the standard deviations as $50\times$ the measure of the grid cells for each LoD, and then exponentially decay their size during training to $5\times$ the measure of the grid cell for each LoD.
This approach proved to be beneficial, particularly at the initial stages of training; allowing for smooth convergence of the Gaussians to regions with high surface density weights.
\section{Experiments}
\label{sec:experiments}
We demonstrate the efficiency of our method in building compact representations with two distinct applications: 2D image fitting~(\cref{sec:image-fitting}), and 3D radiance reconstruction from inverse rendering~(\cref{sec:3D-recon}, \cref{sec:compact}).
We evaluate image reconstruction on four complex high-resolution images, and NeRF reconstruction on the Synthetic blender dataset~\cite{mildenhall2021nerf} and the Tanks \& Temples dataset~\cite{knapitsch2017tanks}.
We conclude by ablating our design choices (\cref{sec:ablation}).

\begin{figure*}[!t]
\centering
\includegraphics[width=\columnwidth]{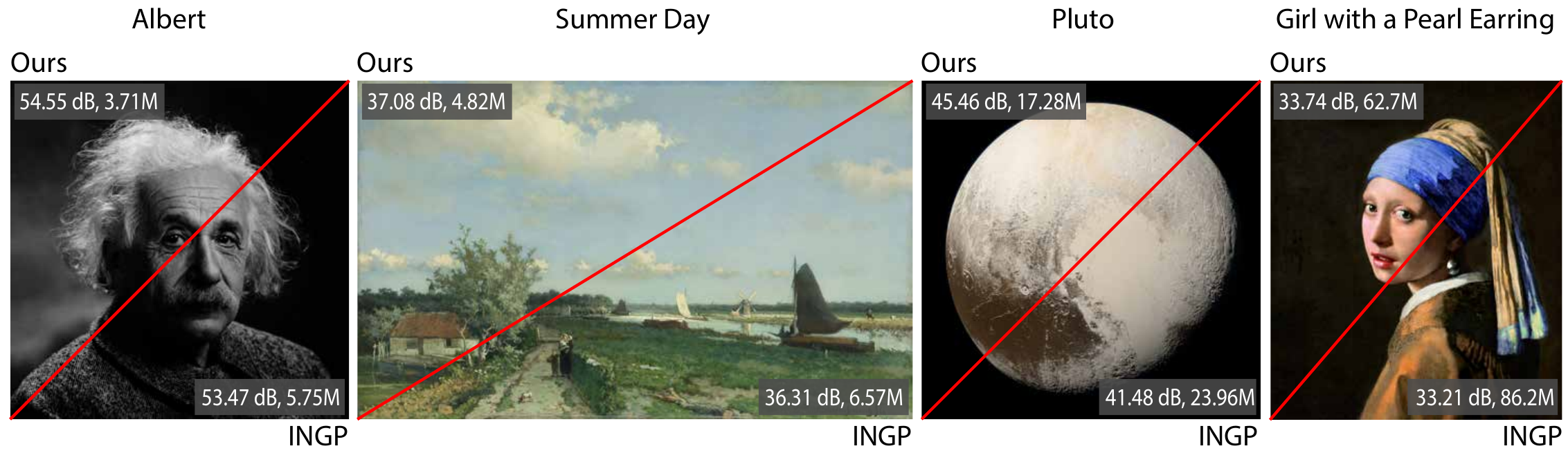}
\caption{
Qualitative comparisons on the giga-pixel images. On each image, we show the reconstruction quality (PSNR) together with the number of parameters.
}
\label{fig:gigapixel}
\end{figure*}

\subsection{2D image fitting (gigapixel)}
\label{sec:image-fitting}
The image fitting task involves learning a mapping between 2D coordinates and image colors, and is a popular benchmark for evaluating neural field methods capabilities in representing high-frequency signals.
We train with an L2 reconstruction loss, and parameterize our models with codebook containing $B=2^{17}$ and $B=2^{18}$ buckets, utilize a maximum of ~$K {=} 4$ Gaussian feature per bucket, and set the maximum resolution $N_{max}$ to half the width of the image.
For this task, the spatial gradient norm of pixel values $\|\nabla I(x_i)\|$ is employed for $W(\x)$ in $\mathcal{L}_{\text{guide}}$ \cref{eq:guide}, hence prioritizing representation of areas within the image containing high-frequency details.

\paragraph{Dataset and baselines.} 
We evaluate our method qualitatively and quantitatively on publicly available gigapixel images, where the total number of pixels ranges from 4 M to 213 M, with InstantNGP as a representative baseline.
We train both our network and InstantNGP with Adam~\cite{kingma2014adam} for $350$ epochs with a learning rate of $1e^{-2}$ and parameters $\beta_1 {=} 0.9$, $\beta_2 {=} 0.99$, $\epsilon {=} 10^{-15}$, while the learning rate for Gaussian positions is set to $1e^{-3}$.
We use a batch size of $2^{16}$.

\begin{table*}[t]
\centering
\caption{We quantitatively compare our method with InstantNGP \cite{muller2022instant} on four giga-pixel images: Girl with a Pearl Earring (Girl), Pluto, Summer Day(Summer), and Albert. 
We compare average PSNR$\uparrow$ and average \texttt{\#} paramaters$\downarrow$.}
\resizebox{0.9\linewidth}{!}{
\setlength{\tabcolsep}{5pt}
\begin{tabular}{cc|cccc|c}
\toprule
Method & \texttt{\#} Params & Girl & Pluto & Summer & Albert & Avg. \\
\midrule
InstantNGP($B=2^{19}$) &  11.91M  & \cellcolor{2nd}28.73 & \cellcolor{2nd}38.59 & \cellcolor{2nd}37.26 & 53.81 & \cellcolor{2nd}39.60 \\
Ours($B=2^{17}$) & \cellcolor{1st}4.56M & 27.60 & 37.53 & 37.08 & \cellcolor{2nd}54.55 & 39.19\\
Ours($B=2^{18}$) & \cellcolor{2nd}8.41M & \cellcolor{1st}28.83 & \cellcolor{1st}39.72 & \cellcolor{1st}38.93 & \cellcolor{1st}55.35 & \cellcolor{1st}40.71 \\
\bottomrule
\end{tabular}
}
\label{tab:gigapixel}
\end{table*}


\paragraph{Discussion.} 
As shown in \Cref{fig:gigapixel} and 
\Cref{tab:gigapixel} our method reconstructs high-fidelity gigapixel images in comparison to InstantNGP while also being a $2.6\times$ more compact representation. 
Note the superior performance on images characterized by localized high-frequency features~(e.g.~Pluto)
This aspect of our method is also highlighted in the pareto plot in \Cref{fig:pareto_plot}, where our method consistently outperforms InstantNGP on different parameters counts.

\subsection{Novel view synthesis}
\label{sec:3D-recon}
We now demonstrate the applicability of our method in a NeRF \cite{mildenhall2021nerf} setup.
Differently from the 2D image fitting task, in the NeRF setting, a volumetric shape is parameterized by a spatial (3D) density function and a spatio-directional~(5D) radiance. 
We demonstrate that our method is capable of solving this problem better than baselines while requiring fewer parameters.

\begin{figure}
\centering
\includegraphics[width=\columnwidth]{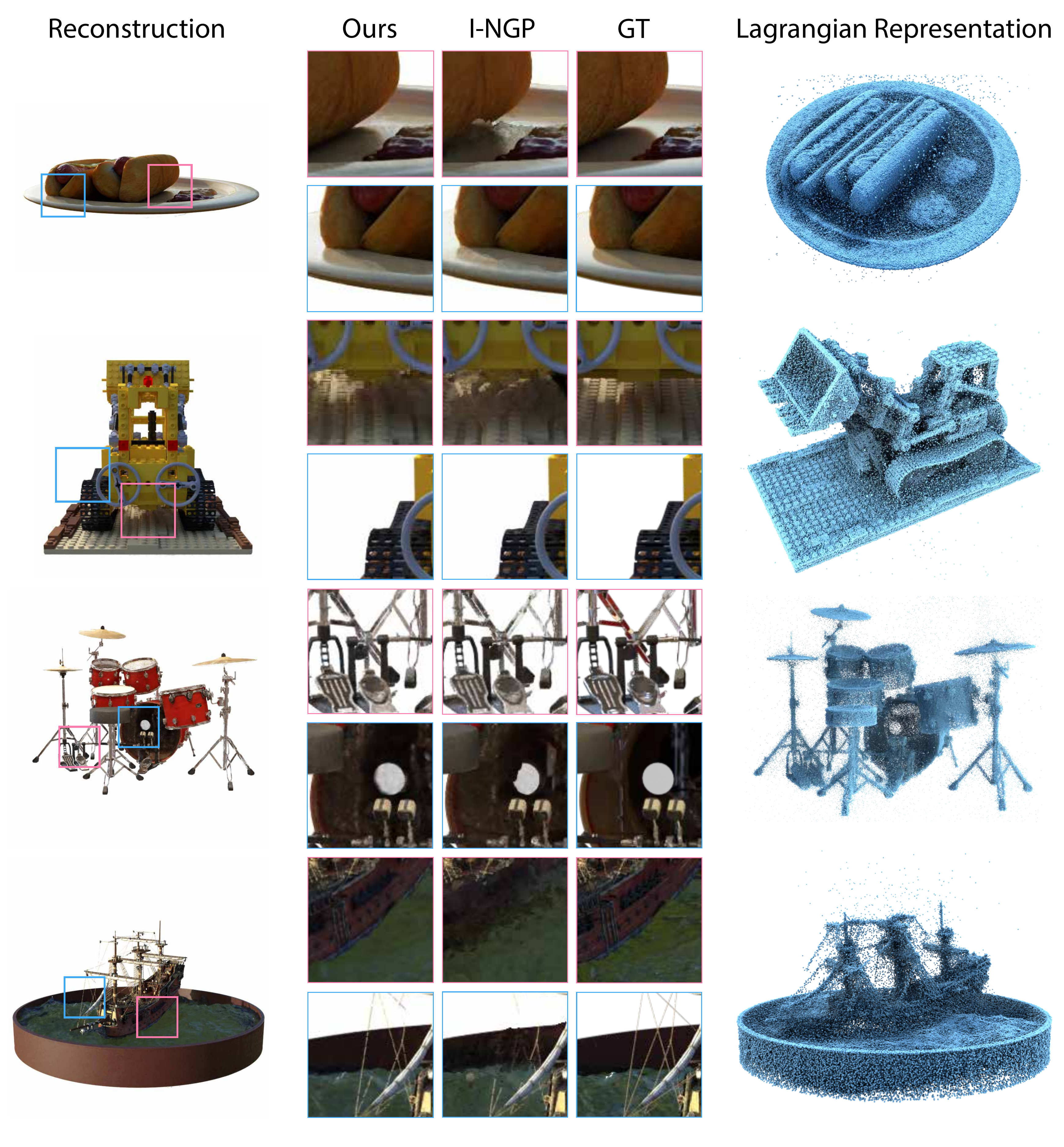}
\caption{Qualitative comparisons on the Synthetic NeRF Dataset \cite{mildenhall2021nerf}. The leftmost column (reconstruction) shows the full-image results of our method and the rightmost column shows the Lagrangian Representation which is learned by our model for the particular scene. }
\label{fig:synthetic_comp}
\end{figure}

\paragraph{Dataset and baselines.} 
We evaluate our method qualitatively and quantitatively with InstantNGP on two widely used benchmarks: the NeRF synthetic dataset \cite{mildenhall2021nerf} and the Tanks \& Temples real-world dataset \cite{knapitsch2017tanks}. 
For Tanks \& Temples, for both our method and InstantNGP, we use mask supervision and train without background modeling.
We evaluate each method both qualitatively~(\Cref{fig:synthetic_comp}), and quantitatively~(\Cref{tab:synth-big,tab:tt-big}), using peak signal-to-noise ratio~(PSNR) as the metric.

\begin{figure}
    \centering
    \includegraphics[width=.95\columnwidth]{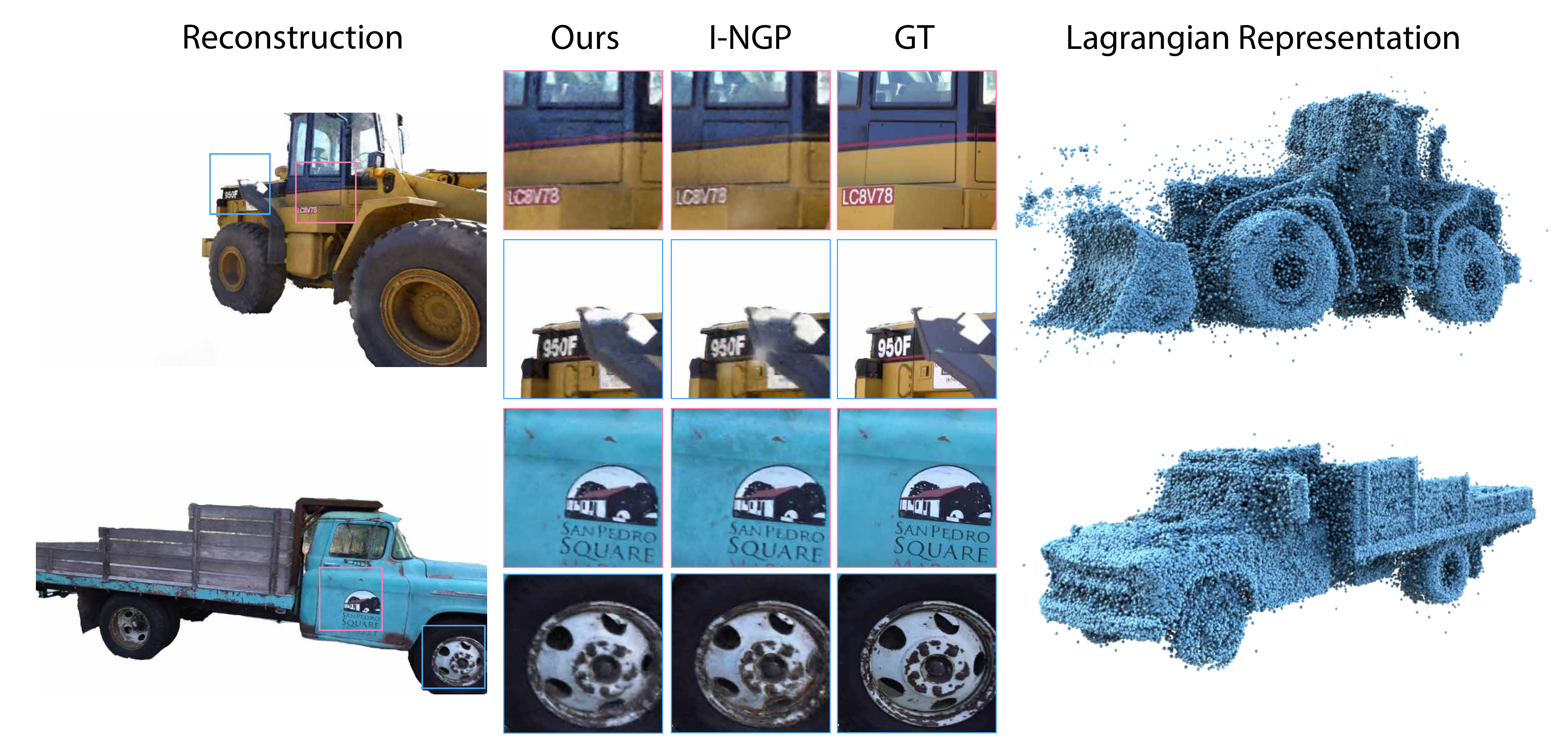}
    \caption{Qualitative comparisons on the Tanks and Temples dataset \cite{knapitsch2017tanks}. The leftmost column (reconstruction) shows the full-image results of our method and the Rightmost column shows the Lagrangian Representation which is learned by our model.}
    \label{fig:tanks_comp}
\end{figure}

\paragraph{Implementation.} 
For this task, we train our models with codebook containing $B{=}2^{17}$ and $B{=}2^{17.9}$~(i.e. we match the number of parameters of InstantNGP) buckets.
We employ $K {=} 4$ Gaussian features per bucket for both datasets. 
We evaluate our method's performance with a maximum grid resolution $N_{max}{=}1024$. 
We fix the weights of distortion loss ($\lambda_{dist}$) to $1e^{-3}$ for the NeRF Synthetic dataset and $1e^-2$ for the Tanks \& Temples dataset, for both InstantNGP and our model.
Additionally, we train our network with a guidance loss weight ($\lambda_{guide}$) of $1e^{-1}$ and a warm-up schedule to allow for the coarse structure of the scene to be learnt first.
We train both the models with an Adam optimizer for $20K$ iterations with a learning rate of $1e^{-2}$ and parameters $\beta_1 {=} 0.9$, $\beta_2 {=} 0.99$, $\epsilon {=} 10^{-15}$.
The learning rate of Gaussian positions is set to $1e^{-3}$.

\begin{table*}[t]
\centering
\caption{We compare our method with InstantNGP on the Nerf Synthetic dataset~\cite{mildenhall2021nerf}.
We report PSNR$\uparrow$ and the parameter count $\downarrow$. 
Our method matches the performances of this state-of-the-art method with considerably fewer parameters.}
\resizebox{\linewidth}{!}{
\setlength{\tabcolsep}{5pt}
\begin{tabular}{cc|cccccccc|c}
\toprule
Method & \texttt{\#} Params & Lego & Mic & Materials & Chair & Hotdog & Ficus & Drums & Ship & Avg. \\
 \midrule
InstantNGP($B=2^{19}$) & \cellcolor{2nd}12.10M  & \cellcolor{2nd}35.67 & \cellcolor{1st}36.85 & 29.60 & \cellcolor{2nd}35.71 & \cellcolor{1st}37.37 & \cellcolor{2nd}33.95 & 25.44 & 30.29 & 33.11 \\
Ours($B=2^{17}$) & \cellcolor{1st}6.68M  & 35.60 & 36.45 & \cellcolor{2nd}29.63 & 35.61 & 37.23 & 33.89 & \cellcolor{2nd}25.67 & \cellcolor{2nd}30.84 & \cellcolor{2nd}33.12 \\
Ours($B=2^{17.9}$) & 12.13M & \cellcolor{1st}35.74 & \cellcolor{2nd}36.78 & \cellcolor{1st}29.66 &  \cellcolor{1st}35.76 & \cellcolor{2nd}37.30 & \cellcolor{1st}34.02 & \cellcolor{1st}25.75 & \cellcolor{1st}31.01 & \cellcolor{1st}33.25 \\
\bottomrule
\end{tabular}
}
\label{tab:synth-big}
\end{table*}
\begin{table*}[t]
\centering
\caption{We quantitatively compare our method with InstantNGP on the Tanks and Temple dataset \cite{knapitsch2017tanks}. 
We compare the PSNR$\uparrow$ and the parameter count $\downarrow$. 
Notably, we are able to match the performances of this state-of-the-art method with considerably fewer parameters. }
\resizebox{0.8\linewidth}{!}{
\setlength{\tabcolsep}{10pt}
\begin{tabular}{cc|cccc|c}
\toprule
Method & \texttt{\#} Params  & Truck & Barn & Family & Caterpillar & Avg. \\
\midrule
InstantNGP($B=2^{19}$) & \cellcolor{2nd}12.10M & \cellcolor{1st}27.42 & 27.10 & \cellcolor{2nd}33.23 & 26.27 & 28.51 \\
Ours($B=2^{17}$) & \cellcolor{1st}6.68M & 27.31 & \cellcolor{2nd}27.36 & 33.22 & \cellcolor{2nd}26.31 & \cellcolor{2nd}28.55 \\
Ours($B=2^{17.9}$) & 12.13M & 27.38 & \cellcolor{1st}27.66 & \cellcolor{1st}33.36 & \cellcolor{1st}26.33 & \cellcolor{1st}28.68 \\
\bottomrule
\end{tabular}
}
\label{tab:tt-big}
\end{table*}

\paragraph{Discussion.}
As shown in \Cref{tab:synth-big,tab:tt-big} our method matches the quality of reconstruction of InstantNGP while achieving a $1.8\times$ more compact representation.
In~\Cref{fig:tanks_comp}, we qualitatively evaluate our method on real data containing complex structures and reflections.
These evaluations highlight our method's superior performance in handling complex scenes.
Notably, our approach demonstrates its capability to resolve collisions that, within the InstantNGP framework, lead to the formation of micro-structures on smooth surfaces (as observed on the truck model, and wheels of the truck). 
In Figure~\ref{fig:synthetic_comp}, we present a visual comparison that highlights the improved rendering of thin structures, such as the legs of a drum instrument, the mast of a ship, and the structure of a hotdog.
Figure~\ref{fig:synthetic_comp} also illustrates our Lagrangian representation, where we demonstrate how the Gaussian in our model converge to regions of high surface density in 3D space.

\subsection{Compact representation}
\label{sec:compact}
We now demonstrate the efficacy of our model in developing compact representation in both Image fitting and NeRF reconstruction applications.
In this experiment, we study decreasing the number of parameters to demonstrate the capability of our method to focus the limited capacity of the codebook towards capturing the high-frequency details of the signal.

\paragraph{Dataset and baselines.}
We evaluate our method quantitatively with CompactNGP \cite{takikawa2023compact}, a compressed variant of InstantNGP, on the NeRF synthetic dataset.
Since the source code for CompactNGP is \ul{not publicly available}, we compare our framework with the scores directly taken from the paper.
We also include in our analysis a comparison with InstantNGP, focusing on the quality of the reconstruction and compactness of the representation, across a range of hyper-parameter configurations on NeRF reconstruction and 2D image fitting.
For plotting our compression graph, we evaluate both our model and InstantNGP on Tanks \& Temples for NeRF reconstruction and gigapixel images fitting.

\begin{table*}[t]
\centering
\caption{We quantitatively compare our method with CompactNGP \cite{takikawa2023compact} on the Nerf Synthetic Dataset \cite{mildenhall2021nerf}. 
We compare the PSNR$\uparrow$ and the parameter count $\downarrow$.
We outperform the recently published CompactNGP while matching in parameter count.}
\resizebox{\linewidth}{!}{
\setlength{\tabcolsep}{10pt}
\begin{tabular}{cc|cccccccc|c}
\toprule
Method & \texttt{\#} Params & Lego & Mic & Materials & Chair & Hotdog & Ficus & Drums & Ship & Avg. \\
\midrule
InstantNGP($B=2^{14}$) & 0.50M & \cellcolor{2nd}32.03 & \cellcolor{1st}35.08 & \cellcolor{2nd}28.73 & \cellcolor{1st}32.59 & \cellcolor{2nd}34.99 & 30.99 & \cellcolor{1st}25.36 & \cellcolor{2nd}27.71 & \cellcolor{1st}30.94 \\
3DGS($\#G=12k$) & 0.60M & 23.92 & 26.05 & \cellcolor{1st}28.87 & 22.72 & 21.43 & 34.43 & 25.26 & 22.13 & 25.60\\
CompactNGP(from \cite{takikawa2023compact}) & \cellcolor{1st}0.18M & \cellcolor{1st}32.31 & \cellcolor{2nd}33.88 & 28.32 & 32.05 & 34.26 & \cellcolor{1st}32.05 & 24.71 & \cellcolor{2nd}27.71 & 30.66 \\
Ours($B=2^{11}$) & \cellcolor{1st}0.18M & 31.15 & 32.65 & 28.52 & \cellcolor{2nd}32.44 & \cellcolor{1st}35.67 & \cellcolor{2nd}31.98 & \cellcolor{2nd}25.07 & \cellcolor{1st}28.26 & \cellcolor{2nd}30.72 \\
\bottomrule
\end{tabular}
}
\label{tab:synth-compact}
\end{table*}

\paragraph{Implementation.}
For these experiments, we train our models with codebook containing $B=2^{11}$ buckets, and we employ a maximum of $K {=} 6$ Gaussian features per bucket and evaluate the model at a maximum grid resolution~$N_{max}{=}256$.
We disable the distortion loss ($\lambda_{dist} {=} 0$) for our network to enable a fair comparison with CompactNGP, that did not use this loss.
Additionally, we train our network with a guidance loss weight~($\lambda_{guide}$) of $1e^{-1}$ and a warm-up scheduler.
Following CompactNGP, we train our model with an Adam optimizer for $35K$ iterations with a learning rate of $1e^{-2}$ and parameters $\beta_1 {=} 0.9$, $\beta_2 {=} 0.99$, $\epsilon {=} 10^{-15}$.
The learning rate of Gaussians is set to $1e^{-3}$.

\begin{figure}[t]
\begin{center}
\includegraphics[width=.495\columnwidth]{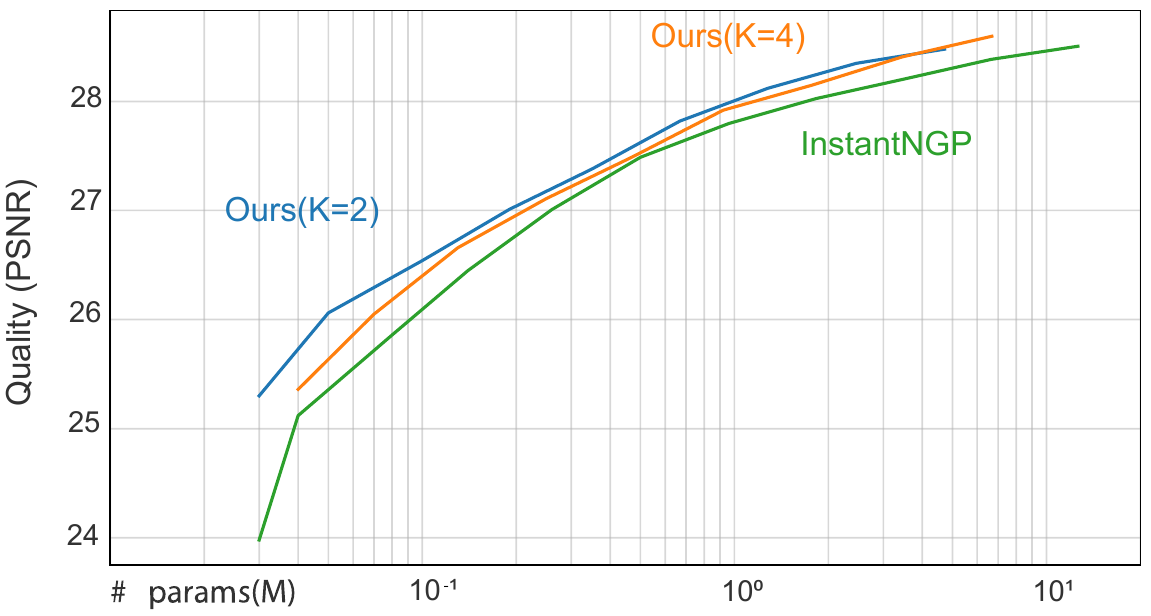}
\hfill
\includegraphics[width=.495\columnwidth]{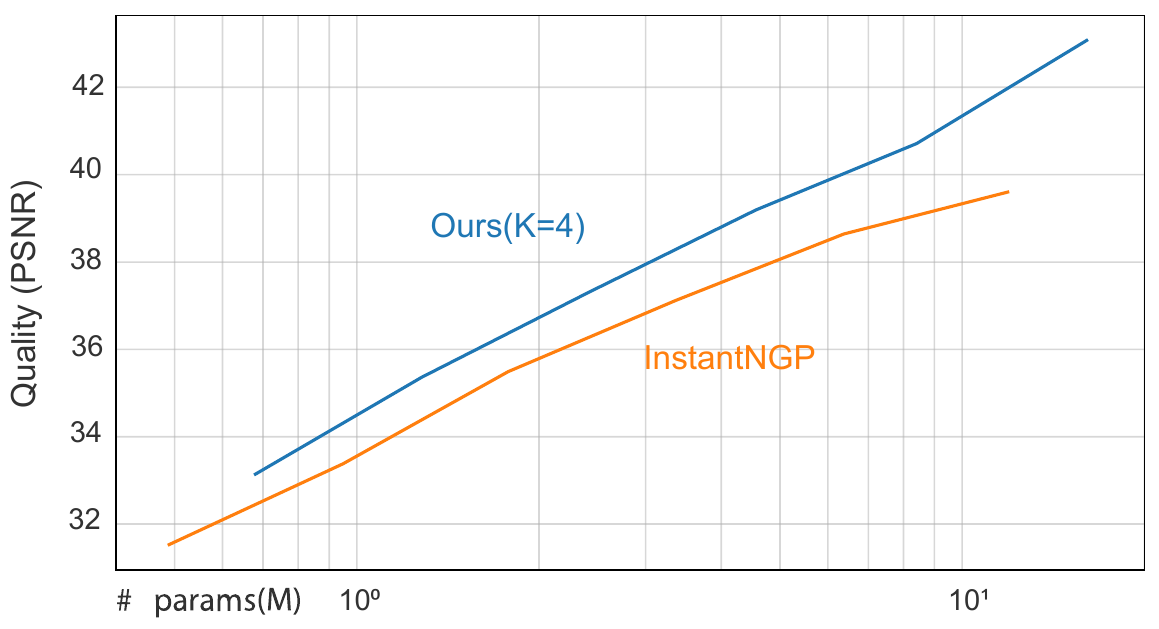}
\end{center}
\caption{
Pareto plot: Tanks and temples(left), Gigapixel images(right). We demonstrate that our method consistently outperforms InstantNGP in terms of quality vs number of parameters.
}
\label{fig:pareto_plot}
\end{figure}

\paragraph{Discussion.}
As shown in \cref{tab:synth-compact}, our method quantitatively outperforms the recently proposed CompactNGP method \ul{on its primary task}, while achieving the same level of compression, a $2.8\times$ more compact representation when compared to InstantNGP.
In \Cref{fig:pareto_plot}, we show that our pareto front lies ahead of InstantNGP in both NeRF application and image fitting, meaning that our model \textit{consistently} outperforms InstantNGP in terms of quality vs. number of parameters.

\subsection{Ablations} 
\label{sec:ablation}
We validate our method in the NeRF reconstruction on the real-world scenes from Tanks \& Temples.
We investigate the importance of the losses, the number of Gaussian features per codebook, and the number of Lagrangian levels. 

\begin{figure}[t]
    \centering
    \begin{subfigure}[c]{0.28\textwidth}
        \centering
        \begin{tabular}{lr}
        \toprule
          & PSNR \\ 
        \midrule
        Full & \textbf{27.94}  \\ 
      $ \text{ w/o } \mathcal{L}_{\text{dist}} $ & 27.70  \\ 
      $ \text{ w/o } \mathcal{L}_{\text{guide}}$ & 27.75  \\ 
        \bottomrule
        \end{tabular}
        \caption{Impact on PSNR} 
        \label{tab:abs_psnr_loss}
    \end{subfigure}
    \hfill
    \begin{subfigure}[c]{0.35\textwidth}
        \centering
        \includegraphics[width=\textwidth]{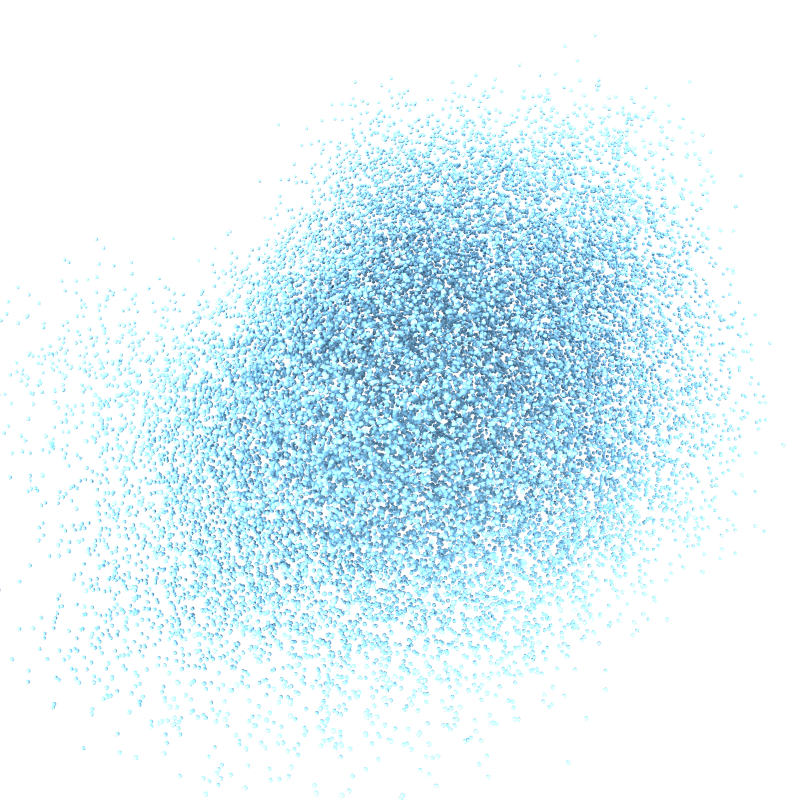}
        \caption{w/o $\mathcal{L}_{\text{guide}}$}
        \label{fig:image1}
    \end{subfigure}
    \hfill
    \begin{subfigure}[c]{0.35\textwidth}
        \centering
        \includegraphics[width=\textwidth]{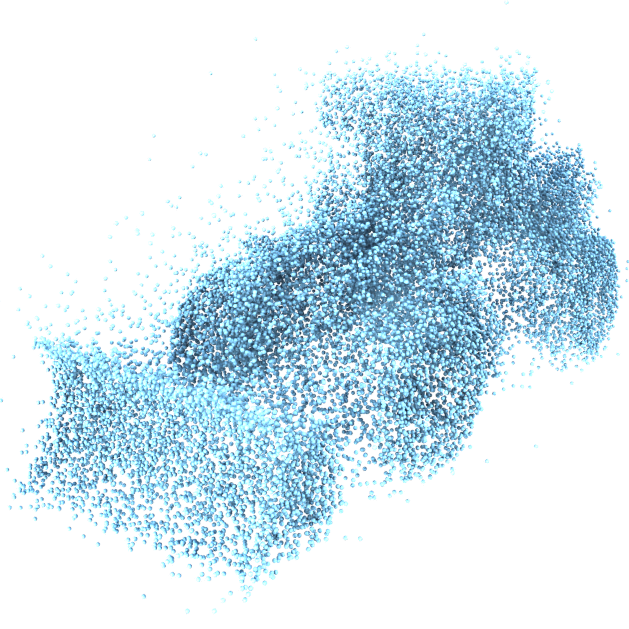}
        \caption{w/ $\mathcal{L}_{\text{guide}}$}
        \label{fig:image2}
    \end{subfigure}
    \caption{
        \textbf{Impact of losses.}
        We show the proposed losses are essential to have the optimal PSNR. We also show that $\loss{guide}$ is critical to place points onto the surface~(i.e., locations in space that need more capacity to be represented), which is the key to achieve a good compression rate.
    }
    \label{fig:abs_loss}
\end{figure}
\begin{table}[t]
\begin{minipage}[t]{0.42\linewidth}
    \centering
    \captionof{table}{
        {\bf Num. of Gaussian ($K$).} 
        We report the storage (the number of parameters) and PSNR with the different number of Gaussian in each bucket. We observe that 4 Gaussians achieve a better trade-off between performance and storage.  
        }
    \centering
    \resizebox{\linewidth}{!}{
        \setlength{\tabcolsep}{10pt}
        \begin{tabular}{lcccc}
        \toprule
        \texttt{\#} of Gauss. & No mixture & 2 & \textbf{4} & 8 \\
        \midrule
        \texttt{\#} Params. (M) & 0.50 & 0.67 & \textbf{0.92} & 1.41 \\
        PSNR & 27.49 & 27.82 & \textbf{27.94} & 27.99 \\
        \bottomrule
        \end{tabular} 
    }
    \label{tab:ablation_numgaussian} 
\end{minipage}
\hfill
\begin{minipage}[t]{0.54\linewidth}
    \centering
    \captionof{table}{
        {\bf Num. of levels ($\cutoffLevel$).} 
        We report the storage (the number of parameters) and PSNR with the different number of Lagrangian levels. We observe that the model of 2 Lagrangian levels at the finest level Gaussians achieves the better trade-off between performance and storage.  
    }
        \centering
        \resizebox{\linewidth}{!}{
            \setlength{\tabcolsep}{10pt}
            \begin{tabular}{lccccc}
            \toprule
            \texttt{\#} of levels & 0 & \textbf{2} & 4 & 8 \\
            \midrule
            \texttt{\#} Params. (M) & 0.5 & \textbf{0.92} & 1.34 & 2.2 \\
            PSNR & 27.49 & \textbf{27.94} & 28.00 & 28.03 \\
            \bottomrule
            \end{tabular} 
        }
        \label{tab:ablation_num_levels} 
\end{minipage}
\end{table}

\paragraph{Loss function -- \cref{fig:abs_loss}.}
We analyze the importance of each loss term except for~$\loss{recon}$.
Note both losses beneficially contribute to the final performance. 
We also qualitatively examine how losses impact point learning, which is a key to learning a compact representation -- by design, we achieve a high compression rate by allocating more points near the object's surface.
We observe that both losses are essential for points, and therefore features, to focus on these surfaces.  
Particularly, $\loss{guide}$
as shown in \cref{fig:abs_loss}, guide random points towards the surface. 
We also observe that $\loss{dist}$, as discussed in MipNeRF360~\cite{barron2022mipnerf360}, eliminates floaters. 

\paragraph{Number of Gaussian features per bucket~($K$) -- \cref{tab:ablation_numgaussian}.}
We study the impact of the number of Gaussians in each bucket on reconstruction performance.
And we show that 4 Gaussians achieve a good trade-off between the number of parameters and performance.  
As shown in ~\cref{tab:ablation_numgaussian}, 8 or even more Gaussians in each bucket, while leading to better performance, start to introduce redundant parameters since the spatial collision is already well addressed with only 4 Gaussians.

\paragraph{Number of Lagrangian levels ($\cutoffLevel$).}
We further investigate the impact of the number of Lagrangian levels.
We observe that two Lagrangian levels in the finest levels where the majority of spatial collisions happen achieve the best trade-off between reconstruction and storage.

\section{Conclusions}
\label{sec:conclusion}
We introduce a neural representation that unifies Eulerian with Lagrangian schools of thought.
We take advantage of the Eulerian grids that allow the use of hash tables, which we extend with the Lagrangian point clouds, imbuing them with the flexibility of point representations.
Taking the best of both worlds, we outperform the state of the art, and provide a better PSNR - parameter count ratio than InstantNGP across various neural field workloads, including real-world scenes.

\newpage
\section*{Acknowledgements}
This work was supported in part by the Natural Sciences and Engineering Research Council of Canada (NSERC) Discovery Grant, NSERC Collaborative Research and Development Grant, Google DeepMind, Digital Research Alliance of Canada, the Advanced Research Computing at the University of British Columbia, Microsoft Azure, and the SFU Visual Computing Research Chair program.

%
%

\begin{thebibliography}{10}
\providecommand{\url}[1]{\texttt{#1}}
\providecommand{\urlprefix}{URL }
\providecommand{\doi}[1]{https://doi.org/#1}

\bibitem{ahmed1974discrete}
Ahmed, N., Natarajan, T., Rao, K.R.: Discrete cosine transform. IEEE transactions on Computers  (1974)

\bibitem{alexa2004point}
Alexa, M., Gross, M., Pauly, M., Pfister, H., Stamminger, M., Zwicker, M.: Point-based computer graphics. In: ACM SIGGRAPH 2004 Course Notes (2004)

\bibitem{aliev2020neural}
Aliev, K.A., Sevastopolsky, A., Kolos, M., Ulyanov, D., Lempitsky, V.: Neural point-based graphics. In: ECCV. Springer (2020)

\bibitem{balsa2014state}
Balsa~Rodr{\'\i}guez, M., Gobbetti, E., Iglesias~Guitian, J.A., Makhinya, M., Marton, F., Pajarola, R., Suter, S.K.: State-of-the-art in compressed gpu-based direct volume rendering. In: Computer Graphics Forum. Wiley Online Library (2014)

\bibitem{barron2022mipnerf360}
Barron, J.T., Mildenhall, B., Verbin, D., Srinivasan, P.P., Hedman, P.: Mip-nerf 360: Unbounded anti-aliased neural radiance fields. CVPR  (2022)

\bibitem{chang2023neural}
Chang, M., Sharma, A., Kaess, M., Lucey, S.: Neural radiance field with lidar maps. In: CVPR (2023)

\bibitem{chen2022tensorf}
Chen, A., Xu, Z., Geiger, A., Yu, J., Su, H.: Tensorf: Tensorial radiance fields. ARXIV  (2022)

\bibitem{chen2024survey}
Chen, G., Wang, W.: A survey on 3d gaussian splatting. ARXIV  (2024)

\bibitem{cheng2020learned}
Cheng, Z., Sun, H., Takeuchi, M., Katto, J.: Learned image compression with discretized gaussian mixture likelihoods and attention modules. In: CVPR (2020)

\bibitem{de2016compression}
De~Queiroz, R.L., Chou, P.A.: Compression of 3d point clouds using a region-adaptive hierarchical transform. IEEE TIP  (2016)

\bibitem{dupont2021coin}
Dupont, E., Goli{\'n}ski, A., Alizadeh, M., Teh, Y.W., Doucet, A.: Coin: Compression with implicit neural representations. ICLR  (2021)

\bibitem{eden1986polynomial}
Eden, M., Unser, M., Leonardi, R.: Polynomial representation of pictures. Signal Processing  (1986)

\bibitem{galic2008image}
Gali{\'c}, I., Weickert, J., Welk, M., Bruhn, A., Belyaev, A., Seidel, H.P.: Image compression with anisotropic diffusion. Journal of Mathematical Imaging and Vision  (2008)

\bibitem{glorot2010understanding}
Glorot, X., Bengio, Y.: Understanding the difficulty of training deep feedforward neural networks. In: Proceedings of the thirteenth international conference on artificial intelligence and statistics. JMLR Workshop and Conference Proceedings (2010)

\bibitem{goyal2001theoretical}
Goyal, V.K.: Theoretical foundations of transform coding. IEEE Signal Processing Magazine  (2001)

\bibitem{gray1998quantization}
Gray, R.M., Neuhoff, D.L.: Quantization. IEEE TIP  (1998)

\bibitem{gross2011point}
Gross, M., Pfister, H.: Point-based graphics. Elsevier (2011)

\bibitem{huffman1952method}
Huffman, D.A.: A method for the construction of minimum-redundancy codes. Proceedings of the IRE  (1952)

\bibitem{kerbl20233d}
Kerbl, B., Kopanas, G., Leimk{\"u}hler, T., Drettakis, G.: 3d gaussian splatting for real-time radiance field rendering. TOG (Proc. of SIGGRAPH)  (2023)

\bibitem{kingma2014adam}
Kingma, D.P., Ba, J.: Adam: A method for stochastic optimization. ARXIV  (2014)

\bibitem{knapitsch2017tanks}
Knapitsch, A., Park, J., Zhou, Q.Y., Koltun, V.: Tanks and temples: Benchmarking large-scale scene reconstruction. TOG  (2017)

\bibitem{kobbelt2004survey}
Kobbelt, L., Botsch, M.: A survey of point-based techniques in computer graphics. Computers \& Graphics  (2004)

\bibitem{kopanas2021point}
Kopanas, G., Philip, J., Leimk{\"u}hler, T., Drettakis, G.: Point-based neural rendering with per-view optimization. In: Computer Graphics Forum. Wiley Online Library (2021)

\bibitem{lassner2021pulsar}
Lassner, C., Zollhofer, M.: Pulsar: Efficient sphere-based neural rendering. In: CVPR (2021)

\bibitem{li2023nerfacc}
Li, R., Gao, H., Tancik, M., Kanazawa, A.: Nerfacc: Efficient sampling accelerates nerfs. ARXIV  (2023)

\bibitem{martel2021acorn}
Martel, J.N., Lindell, D.B., Lin, C.Z., Chan, E.R., Monteiro, M., Wetzstein, G.: Acorn: Adaptive coordinate networks for neural scene representation. ARXIV  (2021)

\bibitem{mildenhall2021nerf}
Mildenhall, B., Srinivasan, P.P., Tancik, M., Barron, J.T., Ramamoorthi, R., Ng, R.: Nerf: Representing scenes as neural radiance fields for view synthesis. Communications of the ACM  (2021)

\bibitem{muller2022instant}
M{\"u}ller, T., Evans, A., Schied, C., Keller, A.: Instant neural graphics primitives with a multiresolution hash encoding. TOG  (2022)

\bibitem{ost2022neural}
Ost, J., Laradji, I., Newell, A., Bahat, Y., Heide, F.: Neural point light fields. In: CVPR (2022)

\bibitem{rakhimov2022npbg++}
Rakhimov, R., Ardelean, A.T., Lempitsky, V., Burnaev, E.: Npbg++: Accelerating neural point-based graphics. In: CVPR (2022)

\bibitem{rho2022masked}
Rho, D., Lee, B., Nam, S., Lee, J.C., Ko, J.H., Park, E.: Masked wavelet representation for compact neural radiance fields. In: CVPR (2023)

\bibitem{song2015vector}
Song, Y., Wang, J., Wei, L.Y., Wang, W.: Vector regression functions for texture compression. TOG  (2015)

\bibitem{sridhar2014real}
Sridhar, S., Rhodin, H., Seidel, H.P., Oulasvirta, A., Theobalt, C.: Real-time hand tracking using a sum of anisotropic gaussians model. In: 2014 2nd International Conference on 3D Vision. IEEE (2014)

\bibitem{stauffer1999adaptive}
Stauffer, C., Grimson, W.E.L.: Adaptive background mixture models for real-time tracking. In: CVPR (1999)

\bibitem{strumpler2022implicit}
Str{\"u}mpler, Y., Postels, J., Yang, R., Gool, L.V., Tombari, F.: Implicit neural representations for image compression. In: Computer Vision--ECCV 2022: 17th European Conference, Tel Aviv, Israel, October 23--27, 2022, Proceedings, Part XXVI. Springer (2022)

\bibitem{sun2023pointnerf++}
Sun, W., Trulls, E., Tseng, Y.C., Sambandam, S., Sharma, G., Tagliasacchi, A., Yi, K.M.: Pointnerf++: A multi-scale, point-based neural radiance field. ARXIV  (2023)

\bibitem{nerfdigest}
Tagliasacchi, A., Mildenhall, B.: Volume rendering digest (for nerf). ARXIV  (2022)

\bibitem{takikawa2022variable}
Takikawa, T., Evans, A., Tremblay, J., M{\"u}ller, T., McGuire, M., Jacobson, A., Fidler, S.: Variable bitrate neural fields. In: ACM SIGGRAPH 2022 Conference Proceedings (2022)

\bibitem{takikawa2021neural}
Takikawa, T., Litalien, J., Yin, K., Kreis, K., Loop, C., Nowrouzezahrai, D., Jacobson, A., McGuire, M., Fidler, S.: Neural geometric level of detail: Real-time rendering with implicit 3d shapes. In: CVPR (2021)

\bibitem{takikawa2023compact}
Takikawa, T., M{\"u}ller, T., Nimier-David, M., Evans, A., Fidler, S., Jacobson, A., Keller, A.: Compact neural graphics primitives with learned hash probing. In: SIGGRAPH Asia 2023 Conference Papers (2023)

\bibitem{KaolinWispLibrary}
Takikawa, T., Perel, O., Tsang, C.F., Loop, C., Litalien, J., Tremblay, J., Fidler, S., Shugrina, M.: Kaolin wisp: A pytorch library and engine for neural fields research. \url{https://github.com/NVIDIAGameWorks/kaolin-wisp} (2022)

\bibitem{tang2018real}
Tang, D., Dou, M., Lincoln, P., Davidson, P., Guo, K., Taylor, J., Fanello, S., Keskin, C., Kowdle, A., Bouaziz, S., et~al.: Real-time compression and streaming of {4D} performances. TOG  (2018)

\bibitem{tang2020deep}
Tang, D., Singh, S., Chou, P.A., Hane, C., Dou, M., Fanello, S., Taylor, J., Davidson, P., Guleryuz, O.G., Zhang, Y., et~al.: Deep implicit volume compression. In: CVPR (2020)

\bibitem{wallac1991jpeg}
Wallace, G.K.: The jpeg still picture compression standard. Commun. ACM  (1991)

\bibitem{wu20234d}
Wu, G., Yi, T., Fang, J., Xie, L., Zhang, X., Wei, W., Liu, W., Tian, Q., Wang, X.: 4d gaussian splatting for real-time dynamic scene rendering. ARXIV  (2023)

\bibitem{neuralfields2021}
Xie, Y., Takikawa, T., Saito, S., Litany, O., Yan, S., Khan, N., Tombari, F., Tompkin, J., sitzmann, V., Sridhar, S.: Neural fields in visual computing and beyond. Computer Graphics Forum  (2022)

\bibitem{xu2022point}
Xu, Q., Xu, Z., Philip, J., Bi, S., Shu, Z., Sunkavalli, K., Neumann, U.: Point-nerf: Point-based neural radiance fields. In: CVPR (2022)

\bibitem{yan2023multi}
Yan, Z., Low, W.F., Chen, Y., Lee, G.H.: Multi-scale 3d gaussian splatting for anti-aliased rendering. ARXIV  (2023)

\bibitem{yang2023deformable}
Yang, Z., Gao, X., Zhou, W., Jiao, S., Zhang, Y., Jin, X.: Deformable 3d gaussians for high-fidelity monocular dynamic scene reconstruction. ARXIV  (2023)

\bibitem{yu2023mip}
Yu, Z., Chen, A., Huang, B., Sattler, T., Geiger, A.: Mip-splatting: Alias-free 3d gaussian splatting. ARXIV  (2023)

\bibitem{zhang2022differentiable}
Zhang, Q., Baek, S.H., Rusinkiewicz, S., Heide, F.: Differentiable point-based radiance fields for efficient view synthesis. In: SIGGRAPH Asia 2022 Conference Papers (2022)

\bibitem{zhang2024papr}
Zhang, Y., Peng, S., Moazeni, A., Li, K.: Papr: Proximity attention point rendering. NeurIPS  (2024)

\bibitem{zhang2023frequency}
Zhang, Y., Huang, X., Ni, B., Li, T., Zhang, W.: Frequency-modulated point cloud rendering with easy editing. In: CVPR (2023)

\end{thebibliography}

\clearpage
\setcounter{section}{5}
\setcounter{figure}{7}
\setcounter{table}{6}

{
    \newpage
    \centering
    \Large
    \textbf{Lagrangian Hashing \\ for Compressed Neural Field Representations}\\
    \vspace{0.5em}Supplementary Material \\
    \vspace{1.0em}
}

We visualize the Lagrangian representation across $\tilde{L}(=2)$ levels, the effect of learnable Gaussian standard deviation, other comparisons with PointNeRF~\cite{xu2022point} and PAPR~\cite{zhang2024papr} and more detailed quantitative results for NeRF reconstruction task. Furthermore, we provide more rendering results in \url{https://theialab.github.io/laghashes}.

\section{Limitations}
Our method is currently restricted to object-centric scenes.
Our approach does not achieve an effective Eulerian-Lagrangian hybrid representation for unbounded scenes when trained with the scene contraction function.
Due to our emphasis on learning a compact representation of 3D objects, we left further exploration of Lagrangian representations for unbounded scenes to future work.

\section{Lagrangian representation across $\tilde{L}(=2)$ levels}
\begin{figure}[h]
\begin{center}

\begin{overpic}[width=\columnwidth]{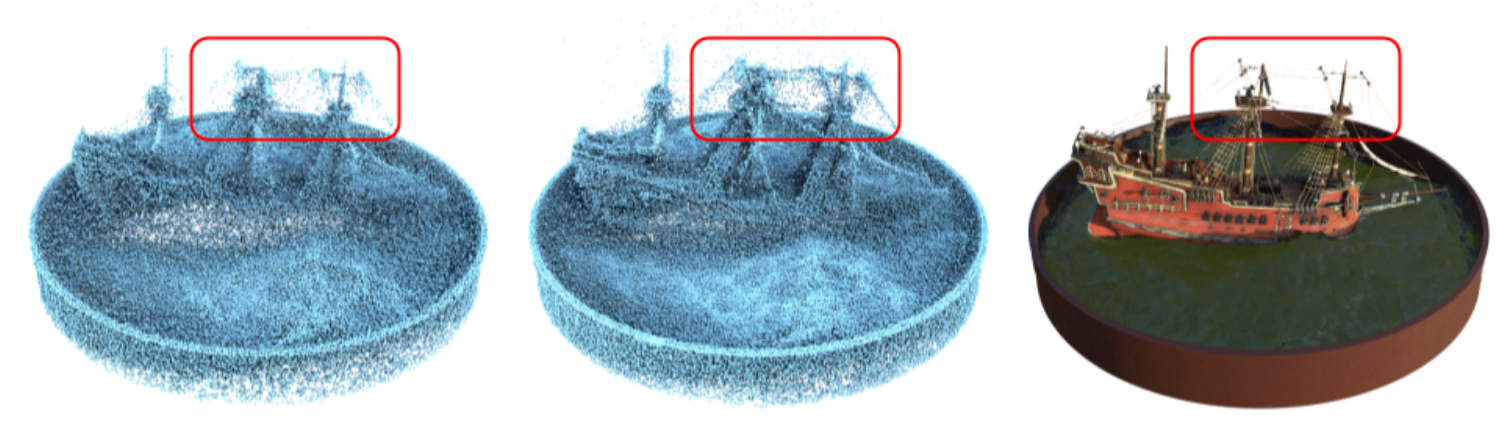}

\put(10, -1){Pre-final LoD}
\put(45, -1){Final LoD}
\put(72.5, -1){Rendered RGB image}
\end{overpic}

\end{center}

\vspace{-1em}
\caption{Notice the mast of the ship(highlighted region), where we see that the final LoD represents high-frequency details better than the pre-final LoD.
}
\label{fig:multi-scale}
\end{figure}

We show our lagrangian representation across $\tilde{L}$ LoDs.
As shown in \cref{fig:multi-scale}, both scales learn the complete point representation and the fine LoD scale tends to capture more high-frequency details.

\begin{table*}[h]
\centering
\caption{\textbf{Learnable vs Fixed standard deviation}: the fixed standard deviation (preferred in our method) is even slightly better than the learnable ones.
}
\resizebox{0.8\linewidth}{!}{
\setlength{\tabcolsep}{10pt}
\begin{tabular}{cc|cccc|c}
\toprule
Method & \texttt{\#} Params  & Truck & Barn & Family & Caterpillar & Avg. \\
\midrule
Fixed $\sigma(B=2^{14})$& 0.92M & 26.77 & 26.75 & 32.30 & 25.94 & 27.92 \\
Trainable $\sigma(B=2^{14})$ & 1.02M & 26.67 & 26.68 & 32.22 & 25.88 & 27.86 \\
\bottomrule
\end{tabular}
}
\label{tab:tt-std}
\end{table*}

\section{Learnable vs fixed standard deviation}
We quantitatively compare the learnable Gaussian standard deviation($\sigma$) vs fixed Gaussian standard deviation($\sigma$).
We show that the fixed standard deviation, simpler and with a smaller number of parameters, achieves a slightly better performance.  
Note this is unsurprising because the fixed standard deviation adaptive to the grid size has the guaranteed spatial coverage whereas the learnable standard deviation might cause degenerate cases (Gaussians that are too big or too small).

\begin{table*}[h]
\centering
\caption{\textbf{Comparisons with other point-based representations, PointNeRF~\cite{xu2022point}, PAPR~\cite{zhang2024papr}, and 3DGS~\cite{kerbl20233d}}: Our method achieves better/equivalent performance (PSNR$\uparrow$) to other representation, even though we don't use point initialization as in~\cite{xu2022point} and don't use 2D U-Net to remove artifacts from the screen space as in~\cite{zhang2024papr}.}
\resizebox{\linewidth}{!}{
\setlength{\tabcolsep}{5pt}
\begin{tabular}{cc|cccccccc|c}
\toprule
Method & \texttt{\#} Params & Lego & Mic & Materials & Chair & Hotdog & Ficus & Drums & Ship & Avg. \\
\midrule
InstantNGP($B=2^{19}$) & 12.10M  & 35.67 & 36.85 & 29.60 & 35.71 & 37.37 & 33.95 & 25.44 & 30.29 & 33.11 \\
3DGS(\#G=210k)~\cite{kerbl20233d} & 12.35M & 35.89 & 36.71 & 30.48 & 35.37 & 38.05 & 35.48 & 26.24 & 31.64 & 33.73 \\
\midrule
PointNeRF~\cite{xu2022point} & \cellcolor{1st}5.00M & 32.65 & 35.54 & 26.97 & \cellcolor{2nd}35.09 & 35.49 & 33.24 & 25.01 & 30.18 & 31.77 \\
PAPR~\cite{zhang2024papr} & 6.80M & 32.62 & 35.64 & \cellcolor{2nd}29.54 & 33.59 & 36.40 & \cellcolor{1st}36.50 & 25.35 & 26.92 & 32.07 \\
3DGS(\#G=110k)~\cite{kerbl20233d} & 6.83M & \cellcolor{2nd}35.28 & \cellcolor{1st}36.54 & \cellcolor{1st}30.46 & 34.80 & \cellcolor{1st}37.77 & \cellcolor{2nd}35.48 & \cellcolor{1st}26.19 & \cellcolor{1st}31.48 & \cellcolor{1st}33.50 \\
Ours($B=2^{17}$) & \cellcolor{2nd}6.68M  & \cellcolor{1st}35.60 & \cellcolor{2nd}36.45 & \cellcolor{2nd}29.63 & \cellcolor{1st}35.61 & \cellcolor{2nd}37.23 & 33.89 & \cellcolor{2nd}25.67 & \cellcolor{2nd}30.84 & \cellcolor{2nd}33.12 \\
\bottomrule
\end{tabular}
}
\vspace{-2em}
\label{tab:psnr-synth-other}
\end{table*}

\begin{table*}[h]
\centering
\caption{\textbf{Comparisons with 3DGS~\cite{kerbl20233d}}: When the parameter count is lowered, our method achieves better performance (PSNR $\uparrow$) while we find that 3DGS experiences a sharp decline in novel view synthesis quality.}
\resizebox{\linewidth}{!}{
\setlength{\tabcolsep}{3pt}
\begin{tabular}{cc|cccccccc|c}
\toprule
Method & \texttt{\#} Params & Lego & Mic & Materials & Chair & Hotdog & Ficus & Drums & Ship & Avg. \\
\midrule
3DGS(\#G=210k)~\cite{kerbl20233d} & 12.35M & 35.89 & 36.71 & 30.48 & 35.37 & 38.05 & 35.48 & 26.24 & 31.64 & 33.73 \\
\midrule
3DGS(\#G=3.5k)~\cite{kerbl20233d} & 0.18M & 16.40 & 23.27 & 15.98 & 19.66 & 18.59 & 28.05 & 15.60 & 21.74 & 19.91 \\
Ours($B=2^{11}$) & 0.18M & \cellcolor{1st}{31.15} & \cellcolor{1st}{32.65} & \cellcolor{1st}{28.52} & \cellcolor{1st}{32.44} & \cellcolor{1st}{35.67} & \cellcolor{1st}{31.98} & \cellcolor{1st}{25.07} & \cellcolor{1st}{28.26} & \cellcolor{1st}{30.72} \\
\bottomrule
\end{tabular}
}
\vspace{-1em}
\label{tab:psnr-3dgs}
\end{table*}
\section{Other comparisons: PointNeRF, PAPR, and 3DGS}
We further show quantitative comparison of our method with other point-based representations, PointNeRF \cite{xu2022point}, PAPR \cite{zhang2024papr}.
As shown in \cref{tab:psnr-synth-other}, our method quantitatively outperforms PointNeRF even though we do not use a COLMAP-based initialization for our point representation.
We also quantitatively outperform PAPR while we learn 3D consistent point representation and do not use the 2D U-Net architecture that PAPR proposed to remove artifacts from the screen space. 
While our method performs similarly to 3DGS at high parameter settings, we observe that 3DGS experiences a significant drop in novel view synthesis quality at lower parameter counts, as shown in \cref{tab:psnr-3dgs}, whereas our method maintains comparatively high quality.

\section{More Detailed Quantitative results}
We provide the metric scores broken down by scene on both datasets. 
Table \ref{tab:synth-ssim} and \ref{tab:synth-lpips} shows the per-scene scores for the NeRF Synthetic dataset and Table \ref{tab:tt-ssim} and \ref{tab:tt-lpips} shows the scores for the Tanks \& Temples dataset.
\begin{table*}[!hbtp]
\centering
\caption{\textbf{NeRF synthetic dataset -- SSIM scores}}
\resizebox{\linewidth}{!}{
\setlength{\tabcolsep}{5pt}
\begin{tabular}{c|cccccccc|c}
\toprule
Method & Lego & Mic & Materials & Chair & Hotdog & Ficus & Drums & Ship & Avg. \\
 \midrule
InstantNGP($B=2^{19}$) & 0.977 & 0.989 & 0.945 & 0.985 & 0.980 & 0.981 & 0.934 & 0.859 & 0.956 \\
Ours($B=2^{17}$) & 0.978 & 0.991 & 0.947 & 0.984 & 0.981 & 0.981 & 0.934 & 0.892 & 0.961\\
\bottomrule
\end{tabular}
}
\label{tab:synth-ssim}
\end{table*}

\begin{table*}[!hbtp]
\caption{\textbf{NeRF synthetic dataset -- LPIPS scores}}
\centering
\resizebox{\linewidth}{!}{
\setlength{\tabcolsep}{5pt}
\begin{tabular}{c|cccccccc|c}
\toprule
Method & Lego & Mic & Materials & Chair & Hotdog & Ficus & Drums & Ship & Avg. \\
 \midrule
InstantNGP($B=2^{19}$) & 0.027 & 0.017 & 0.072 & 0.028 & 0.037 & 0.038 & 0.086 & 0.180 & 0.065 \\
Ours($B=2^{17}$) & 0.027 & 0.015 & 0.070 & 0.024 & 0.036 & 0.049 & 0.083 & 0.139 & 0.055\\
\bottomrule
\end{tabular}
}
\label{tab:synth-lpips}
\end{table*}
\begin{table*}[!hbtp]
\centering
\caption{\textbf{Tanks \& Temples - SSIM scores}}
\resizebox{0.8\linewidth}{!}{
\setlength{\tabcolsep}{10pt}
\begin{tabular}{c|cccc|c}
\toprule
Method & Truck & Barn & Family & Caterpillar & Avg. \\
\midrule
InstantNGP($B=2^{19}$) & 0.913 & 0.837 & 0.955 & 0.912 & 0.904 \\
Ours($B=2^{17}$) & 0.910 & 0.848 & 0.954 & 0.910 & 0.906 \\
\bottomrule
\end{tabular}
}
\label{tab:tt-ssim}
\end{table*}

\begin{table*}[!hbtp]
\caption{\textbf{Tanks \& Temples - LPIPS scores}}
\centering
\resizebox{0.8\linewidth}{!}{
\setlength{\tabcolsep}{10pt}
\begin{tabular}{c|cccc|c}
\toprule
Method & Truck & Barn & Family & Caterpillar & Avg. \\
\midrule
InstantNGP($B=2^{19}$) & 0.138 & 0.289 & 0.076 & 0.152 & 0.164 \\
Ours($B=2^{17}$) & 0.144 & 0.280 & 0.079 & 0.153 & 0.164 \\
\bottomrule
\end{tabular}
}
\label{tab:tt-lpips}
\end{table*}

\end{document}